\documentclass[letterpaper, 10 pt, conference]{ieeeconf}  

\IEEEoverridecommandlockouts    

\usepackage{graphics} 
\usepackage{epsfig} 
\usepackage{epstopdf}
\usepackage{forest}
\usepackage{lipsum}

\usepackage{graphicx}
\usepackage{siunitx}
\usepackage{multirow}

\usepackage{url}

\usepackage{tikz}
\usepackage{textcomp}
\usepackage{lipsum}

\newcommand\copyrighttext{%
  \footnotesize \textcopyright 2020 IEEE. Personal use of this material is permitted. Permission from IEEE must be obtained for all other uses, in any current or future media, including reprinting/republishing this material for advertising or promotional purposes, creating new collective works, for resale or redistribution to servers or lists, or reuse of any copyrighted component of this work in other works.}
\newcommand\copyrightnotice{%
\begin{tikzpicture}[remember picture,overlay]
\node[anchor=south,yshift=10pt,xshift=10pt] at (current page.south) {\fbox{\parbox{\dimexpr\textwidth-\fboxsep-\fboxrule\relax}{\copyrighttext}}};
\end{tikzpicture}%
}

\title{\LARGE \bf The PREVENTION Challenge: How Good Are Humans Predicting Lane Changes?}

\author{A. Quintanar, R. Izquierdo, I. Parra, D. Fern\'andez-Llorca, and M. A. Sotelo
\thanks{A. Quintanar, R. Izquierdo, I. Parra, D. Fern\'andez-Llorca and M. A. Sotelo are with the Computer Engineering Department, Universidad de Alcal\'a, Alcal\'a de Henares, Spain.
        {\tt\small alvaro.quintanar@uah.es}}%
}

\begin{document}

\maketitle

\copyrightnotice

\thispagestyle{empty}
\pagestyle{empty}

\begin{abstract}
While driving on highways, every driver tries to be aware of the behavior of surrounding vehicles, including possible emergency braking, evasive maneuvers trying to avoid obstacles, unexpected lane changes, or other emergencies that could lead to an accident. In this paper, human's ability to predict lane changes in highway scenarios is analyzed through the use of video sequences extracted from the PREVENTION dataset, a database focused on the development of research on vehicle intention and trajectory prediction. Thus, users had to indicate the moment at which they considered that a lane change maneuver was taking place in a target vehicle, subsequently indicating its direction: left or right. The results retrieved have been carefully analyzed and compared to ground truth labels, evaluating statistical models to understand whether humans can actually predict. The study has revealed that most participants are unable to anticipate lane-change maneuvers, detecting them after they have started. These results might serve as a baseline for AI's prediction ability evaluation, grading if those systems can outperform human skills by analyzing hidden cues that seem unnoticed, improving the detection time, and even anticipating maneuvers in some cases.


\end{abstract}

\section{Introduction and Related Work} \label{sec:introduction}

According to \cite{EuropeanComission2018}, about 8\% of road fatalities in the EU occurred on highways, which represents more than 2k people involved. This number must be reduced in upcoming years to meet the proposed objectives \cite{EuropeanCommision2019}, trying to halve the number of serious injuries.
Lane change accidents represent one of the leading causes of motorway accidents. A lane change accident occurs when a vehicle maneuvers laterally from one lane into another, colliding with a vehicle that was already in the destination lane earlier. Nowadays, many manufacturers develop driving aid systems that allow, for example, to determine whether a vehicle is in the blind spot or not, avoiding a possible emergency. Due to the forced coexistence of vehicles equipped with these systems and those lacking them, there is a need to conduct a study that analyses whether humans are capable of detecting and avoiding a lane change accident in advance. 

The use of turn signal is another primary cue used to infer whether a driver desires to start a lane-change maneuver. As stated in \cite{Suzanne2004}, the use of turn signals may vary significantly among drivers, and evaluating data on turn signal could lead to improve prediction schemes.
In some cases, drivers that use turn signals activate them after beginning the lane-change maneuver, as shown in \cite{Hetrick1997}. Our study carefully assesses the lane-change data contained in The PREVENTION Dataset \cite{prevention_dataset}, which includes examples of most common types of maneuvers, including when a driver changes lanes to pass a slower lead vehicle to maintain current speed, a typical situation in Spanish highways.

To detect lane changes, drivers should identify areas of risk approaching and focus on them, e.g., a merge action in a busy highway. Watching for subtle changes in other vehicles' motion to identify the drivers' intent, such as drifting in the lane or slight angling of the front wheels, could help drivers avoiding critical situations. 
A vehicle driver will predict the intentions of other traffic participants to decide to anticipate and evade a potentially dangerous conflict. Thus, correct predictions ease a quick reaction, but incorrect ones could be potentially hazardous. 

According to \cite{Nilsson2016}, a lane-change maneuver is desirable if it is either discretionary, anticipatory, or mandatory (initialized by road conditions). The sequences included in the study display different lane-change maneuvers, including double-triple simultaneous LCs, hazardous maneuvers, emergency vehicles in traffic jams and other circumstances.

\begin{figure}[t]
    \centering
    {
    \vspace{2pt}
    \includegraphics[width=\columnwidth]{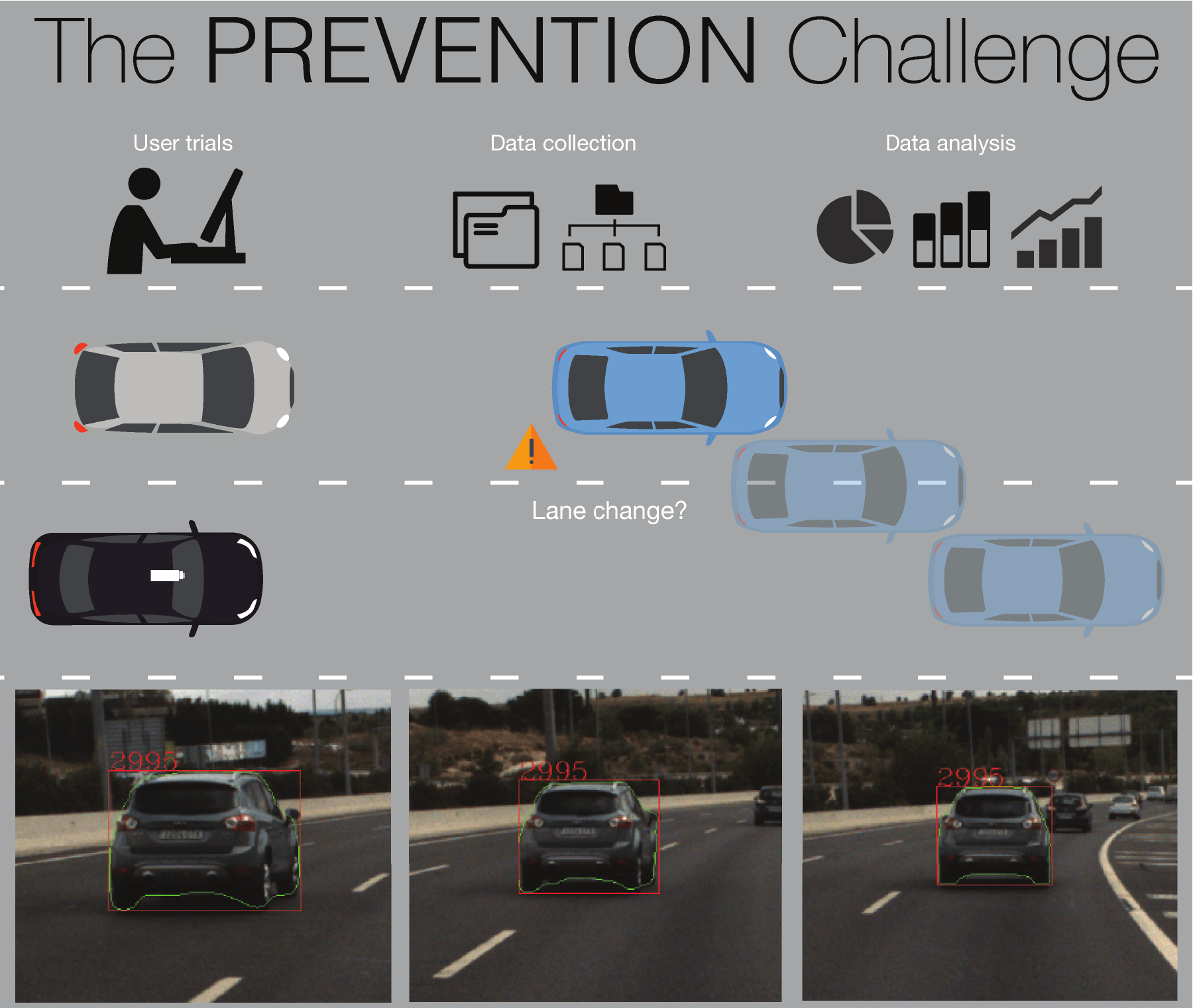}
    }
    \vspace{-18pt}
    \caption{PREVENTION Challenge overview.}
    \vspace{-16pt}
    \label{fig:intro}
\end{figure}

\subsection{Situational awareness and scene understanding}
Situational awareness (SA) could be defined as the ability to scan the environment and perceive danger, challenges, and opportunities while maintaining the ability to conduct normal activities. Driving in traffic consists of recognizing and reacting to potentially dangerous scenarios, maneuvering efficiently through traffic, and achieving higher-level objectives while addressing other concerns \cite{Sukthankar1994}.

By receiving advanced driver training \cite{Walker2009}, drivers can develop and improve their SA, but not necessarily in the way that is intended. Thus, it is clear that a holistic method to identify and understand every cue present in a specific situation can help drivers to react in those extreme conditions where an accident is near to happen.

In \cite{Zhao2017}, lane changes are classified into five categories regarding the interaction style observed with surrounding vehicles. Most lane-change maneuvers analyzed in this study were self-motivated, and no significant interaction with other vehicles was observed. SA is necessary to understand and react in complex situations, as several vehicles changing lanes at the same time or a congested highway with a vehicle fast approaching that wants to overtake. 

Regarding driving skills of novice drivers, in \cite{DeCraen2011} young drivers seemed to recognize that they were not as skilled as the average driver . Moreover, overconfident drivers were worse at adapting their behavior to the traffic situation, acting more aggressively.
Comparing young and old drivers' behavior in \cite{Romoser} result in the need for old drivers to check their surroundings before performing a lane change. In a dynamic driving, the environment is critical to detect and perceive cues and analyzing the future state of the driving agents based on those cues to react appropriately and anticipate hazardous maneuvers. For old drivers, the post-drive feedback was useful in terms of increasing their situational awareness.

\subsection{Vehicle intention prediction}

Apart from analyzing the scene by itself, in this study, the \textit{holistic image} of every user is assessed by comparing the driving situation understanding capabilities, predicting other vehicles' lane changes in a great set of sequences. As in \cite{Ohn-Bar2014}, overtake-late (lane crossing) and overtake-early (initial lateral motion) maneuvers are compared with the very moment when the user considers that a lane-change maneuver is starting.

In this paper, the PREVENTION dataset \cite{prevention_dataset} has been used to assess human capabilities in detecting and predicting lane changes in highway scenarios. The dataset offers more than 356 minutes, 4 million vehicle detections, and 3000 trajectories, collected from up to 8 sensors, including LiDAR, RADAR, DGNSS, and cameras, acquiring information from surrounding vehicles up to a range of 100 meters.
Among other data, frontal camera information along with both manual and CNN detections, as well as lane-change annotations have been used to create the Challenge. It is essential to notice that participants assessed lane-change maneuvers of surrounding vehicles, not of the ego vehicle. The main objective of this work is to prove if participants are able to anticipate lane-change maneuvers or not.

The paper is organized as follows: Section \ref{sec:methodology} offers an exhaustive description of the methodology applied to create the PREVENTION Challenge, by selecting and adapting the sequences to perform the trials, as well as developing a questionnaire to assess users' demographic information regarding the survey. In Section \ref{sec:results}, several results are described, creating a regression model and analyzing if humans are able to predict lane changes in highway scenarios. Section \ref{sec:discussion} offers a discussion about the results and the Challenge itself, by evaluating the content, trials, and the contribution to the state of the art. Finally, Section \ref{sec:conclusions} completes the paper, providing conclusions extracted from the Challenge and suggesting some future developments of prediction of vehicle intention.

\section{Methodology} \label{sec:methodology}

This section describes the methodology employed to perform the survey, starting by selecting the participants, designing the architecture, revising and fine-tuning the materials used and the process of trial, post-processing the data, and finishing by extracting some possible predictive factors and statistics about human capabilities compared with previously labeled data.

\subsection{Participants}

Participants were recruited among the Engineering School, from students to teachers and other research staff, as well as family, friends, and colleagues. Thus, more significant variability is achieved in terms of the expertise of the participants.

\subsection{Architecture}
\begin{figure}[h!]
    \centering
    {
    \vspace{4pt}
    \includegraphics[width=\columnwidth]{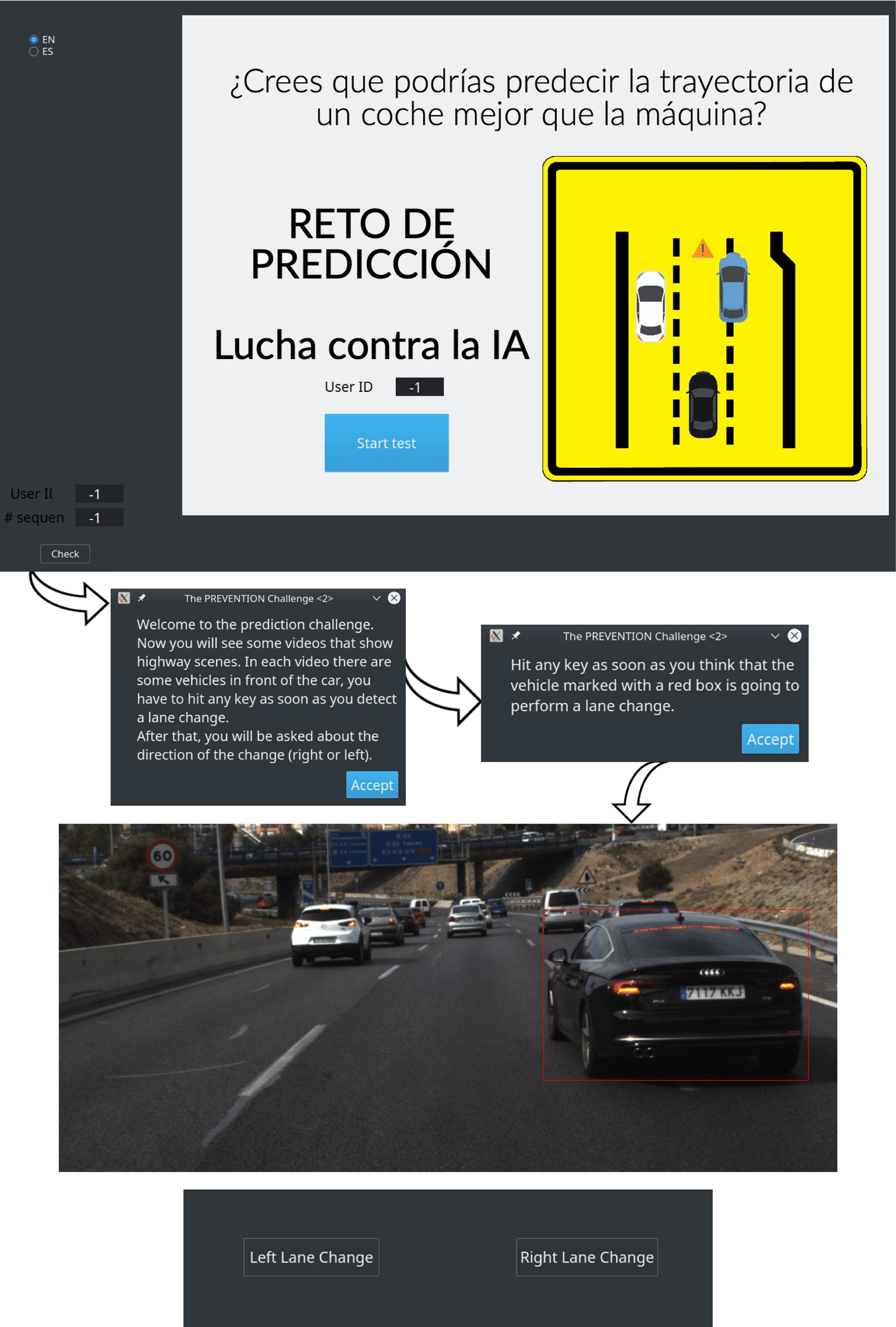}
    }
    \vspace{-18pt}
    \caption{Application outline.}
    \vspace{-18pt}
    \label{fig:app}
\end{figure}

The study was performed using a Qt-based application, where 30 randomly selected sequences were presented to the participants to select the starting instant of a lane change or do nothing in its absence. The vehicle of interest was inside a red bounding box; participants were asked to hit any key as soon as they notice an upcoming lane change maneuver performed by that vehicle. The independent variable was defined as the lane change existence and its direction (three possible options: left lane change, right lane change, or none). If the participant did not spot any lane change, the sequence would continue up to its finishing point, and that answer was marked as a ``none''. The dependent variables were defined as prediction correctness and prediction certainty. Some insights about the application can be observed in Figure \ref{fig:app}.

\subsection{Materials: revision and fine-tuning}

A total of 794 different sequences were selected from the PREVENTION dataset. The dataset offers different sequences recorded in various situations and locations, with precise labeling of lane changes. The sequences are video fragments taken from the front camera, recorded at $\sim$10 Hz, and conveniently cropped to 1920x600, removing the top and bottom bands that contain irrelevant information. Recordings were made in a naturalistic way, driving in real traffic during central hours, avoiding heavy road traffic. 
As the lane changes are already labeled, including blinker use during the maneuver, all lane-change maneuvers were carefully revised, and those that seemed to be unclear for humans to understand, such as far vehicles or ambiguous IDs, were conveniently removed from the trial set. The total amount of LC maneuvers used in the trials and their nature are described in Table \ref{table:ManeuversCount}. 
Nevertheless, the survey also needed scenes where there were vehicles that keep their trajectory in the lane, that is, sequences without lane-change maneuvers of a specific vehicle. To do this task, the authors revised every recording, labeling appropriate fragments of different vehicles, obtaining the sequences of no-lane-change maneuvers described in Table \ref{table:ManeuversCount}.
 
 \begin{table}[h!]
 \vspace{-10pt}
\caption{Maneuvers count\hspace{\textwidth}$^a$ No Lane Change, $^b$ Left Lane Change, $^c$ Right Lane Change}
\label{table:ManeuversCount}
\begin{center}
\vspace{-14pt}
\begin{tabular}{c|c|c|c|c|c}
\textbf{Record \#} & \multirow{2}{*}{\textbf{NLC$^a$}} & \multicolumn{2}{c|}{\textbf{LLC$^b$}} & \multicolumn{2}{c}{\textbf{RLC$^c$}} \\ 

 \textbf{Drive \#}              &              & $\overline{blinker}$   & \textit{blinker}   & $\overline{blinker}$    & \textit{blinker} \\ \hline
\textbf{1 - 1}                  & 11           & 4           & 15           & 7   & 38\\ \hline
\textbf{2 - 1}                  & 7            & 1           & 11           & 4  &  15\\ 
\textbf{2 - 2}                  & 13           & 0           & 13           & 2 &   18\\ \hline
\textbf{3 - 1}                  & 14           & 2           & 12           & 3  &  11\\ 
\textbf{3 - 2}                  & 14           & 2            & 7           & 0  &  19\\ \hline
\textbf{4 - 1}                  & 29           & 5           & 23           & 17  &  59\\ 
\textbf{4 - 2}                  & 24           & 9           & 39           & 15  &  42\\ 
\textbf{4 - 3}                  & 7            & 1            & 7           & 1  &  9\\ \hline
\textbf{5 - 1}                  & 20           & 6           & 24           & 12  &  46\\
\textbf{5 - 2}                  & 20           & 3           & 35           & 9 &   39\\
\textbf{5 - 3}                  & 20           & 1           & 17           & 1 &   11\\ \hline
\multirow{2}{*}{\textbf{Total}}& \multirow{2}{*}{\textbf{179}} & \textbf{34} & \textbf{203}   & \textbf{71}  & \textbf{307}\\ \cline{3-6}
            &   & \multicolumn{2}{c|}{\textbf{237}}  & \multicolumn{2}{c}{\textbf{378}}  \\ \hline
\multirow{2}{*}{\textbf{\%}}& \multirow{2}{*}{\textbf{22.54\%}} & \textbf{4.28\%} & \textbf{25.57\%}   & \textbf{8.94\%}  & \textbf{38.66\%}\\ \cline{3-6}
            &   & \multicolumn{2}{c|}{\textbf{29.85\%}}  & \multicolumn{2}{c}{\textbf{47.61\%}}  \\ \hline
\end{tabular}
\end{center}
\vspace{-5pt}
\end{table}
\normalsize

\begin{figure}[h!]
    \centering
    {
    \includegraphics[width=\columnwidth]{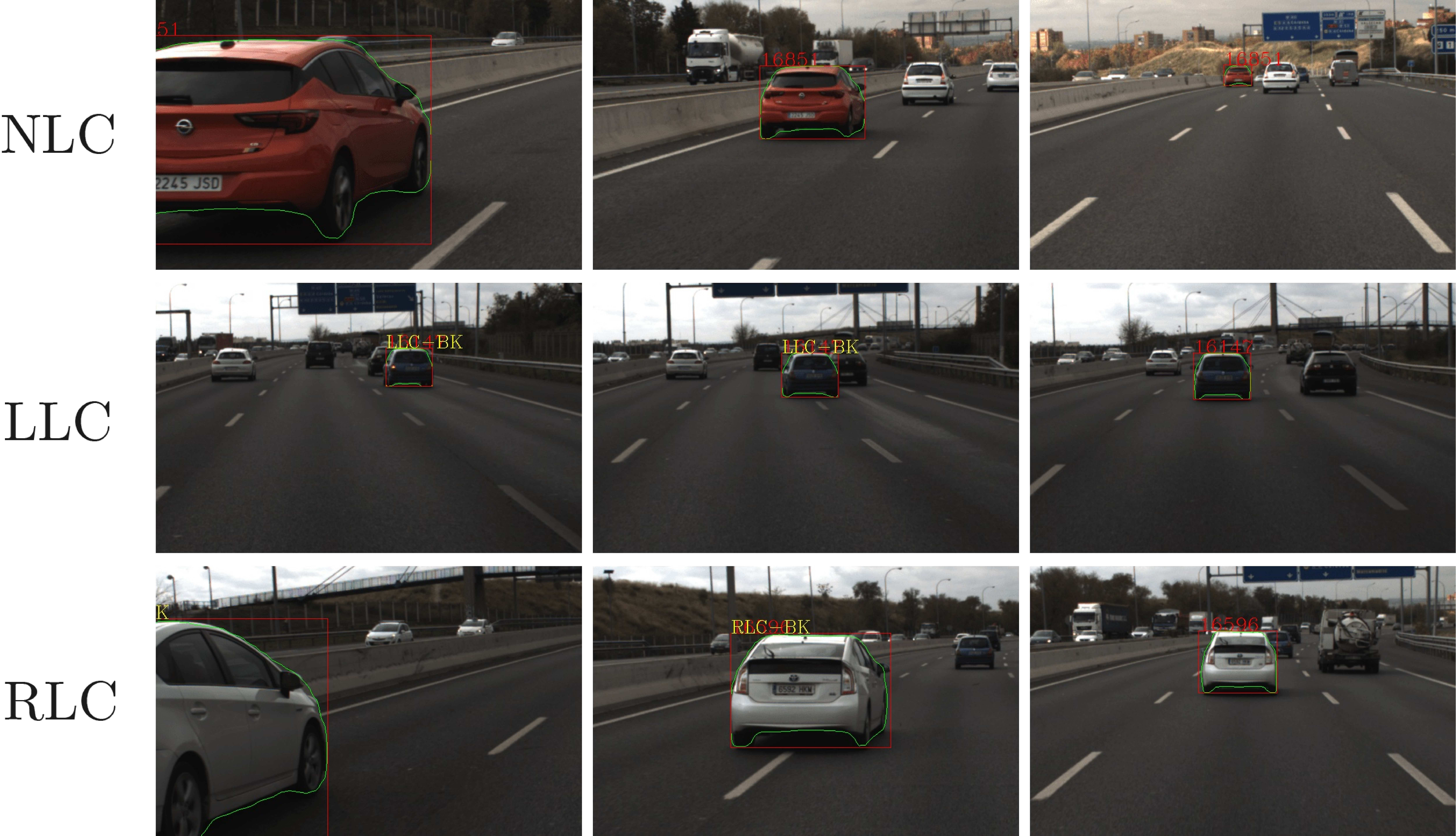}
    }
    \vspace{-18pt}
    \caption{Main frames of various sequences.}
    \vspace{-13pt}
    \label{fig:sequences}
\end{figure}

As there were up to 800 sequences in total, it was necessary to select a few of them for each trial randomly. Thus, all lane change annotations were stacked, filtered, and shuffled randomly, selecting 10 of each type to form the questions for a single participant. To avoid showing the same video of a sequence (to increase variability), a random number of previous frames (between 50 and 100) are showed. Balance of maneuvers type and their selection is discussed in Section \ref{sec:trialContentDiscussion}; in Figure \ref{fig:sequences} various sequences examples are shown.

Regarding making the participant focus on the selected vehicle, firstly, high-level labeling was developed, including the reference of the lane with respect to the ego and the color of the vehicle of interest. After a few test volunteers, that idea was deprecated, and another method based on the bounding box was selected: the vehicle of interest is marked with a red bounding box with a thickness of one pixel, leaving the scene almost unaltered to identify a possible cue that would make the user infer a lane-change maneuver.

\subsection{Collection of demographic information}

Participants were asked to provide demographic details, such as age, gender, occupation, and whether they have a driving license. If so, they were asked additionally about their experience (less than a year, between one and two years or more than two years), their driving frequency (daily, only during weekends or occasionally) and their typical driving zones (urban scenarios or mostly highways and two-way roads). Moreover, optionally they could input their full name and e-mail address if they wished to be notified about the study results.
The question about driving license is crucial to distinguish between actual drivers that could recognize typical traffic situations and those who not, including graduation regarding their experience, driving frequency, and driving scenarios. For example, frequent and experienced drivers who are familiar with highway scenarios will likely predict a merging maneuver before a participant who does not have a driving license.

\subsection{Trials}

A fixed procedure was developed to proceed with every volunteer:
\begin{enumerate}
    \item After filling up the demographics form, together with the user, a single ID is issued, which is going to be introduced in the Challenge application, as it is shown in figure \ref{fig:app}.
    \item After clicking ``Start Test'', a prompt window is showed, explaining the process to the user. The participant is asked to tell the staff what is needed to do, as well as if there is any question.
    \item The prompt window previous to every sequence appears, waiting for the user to click ``Accept'' in order to start the first sequence.
    \item The sequence is displayed, with the vehicle of interest being focused by a red bounding box. If the user detects an upcoming lane-change maneuver, any key will be hit (the volunteer is encouraged to use the space bar).
    \item The sequence stops, and the lane change selection screen is showed, formed by two buttons: ``Left Lane Change`` and ``Right Lane Change''. The user clicks on the selected option.
    \item The prompt window previous to every sequence is showed, asking the user to accept to start the next sequence.
    \item When all sequences have been finished, a results window is showed, featuring:
    \begin{itemize}
        \item Accuracy (\%).
        \item Delay in detecting the lane change (s).
        \item Lane change anticipation (s).
    \end{itemize}
    \item The start window is showed, waiting for a new user ID to start a test.
\end{enumerate}

The Challenge is supposed to be done in a controlled environment, so the staff used laptops with the same screen resolution and controllers for every test. They were instructed to follow the process meticulously, making sure that the participant has understood every step and the purpose of the test. In order to avoid unnecessary waiting times, a parallel thread is used to load the frames of the sequence, which leads to smoother user experience.

\subsubsection{Labeling method}

Lane-change maneuvers are labeled in the dataset as follows:
\begin{itemize}
    \item \texttt{f0}: the frame where the driver starts the maneuver, either activating the turn signal or by showing changes in the trajectory of the vehicle.  These labels have been double-checked to assure that they are correct, representing the initial point of the lane change.
    \item \texttt{f1}: the frame where the rear middle part of the vehicle is just between the lanes; at this point, the lane-change maneuver is virtually ended from the drivers' perception.
    \item \texttt{f2}: the frame where the vehicle is positioned in the new lane, the lane-change maneuver is completed.
\end{itemize}

\subsection{Data analysis}

The result of the test of every participant is coded in a text file, in order to make post-processing and analysis easier. Among others, ground truth annotations for every lane-change step, type of maneuver, and test results (frame selected by the user and kind of maneuver) are included. Thus, the post-processing stage consisted of extracting both data from the test application and demographics form, performing the statistical model creation, including histogram and curve fitting with MATLAB.
\vspace{-9pt}

\section{Results}\label{sec:results}
\vspace{-2pt}
The Challenge was conducted over several weeks, starting in January 2020. A total of 72 participants were tested, assessing 30 randomly selected sequences of highway scenarios. This section presents different results regarding prediction success, regression model, cue analysis, and individual factors as to anticipation of lane-change maneuvers. 

\subsection{Prediction success}

The dataset used was recorded around Madrid, in the center of Spain, so, logically, local users that maybe commute daily to get to their work or studies could be familiar with them. Some of them stated that during the trial process, being curious about the recording areas.
After processing the results, we find it appropriate to remove the worst and the best trial of each participant to avoid biasing or random variation. The measures of delay and anticipation only include hits (that is, when the user response is the same as the ground truth), so the accuracy and missed maneuvers are analyzed in this section as well. Anticipation and delay are converted from frames to seconds, taking into account that the sequences are recorded at $\sim$10 Hz.

\begin{figure}[h!]
    \centering
    {
    \includegraphics[width=\columnwidth]{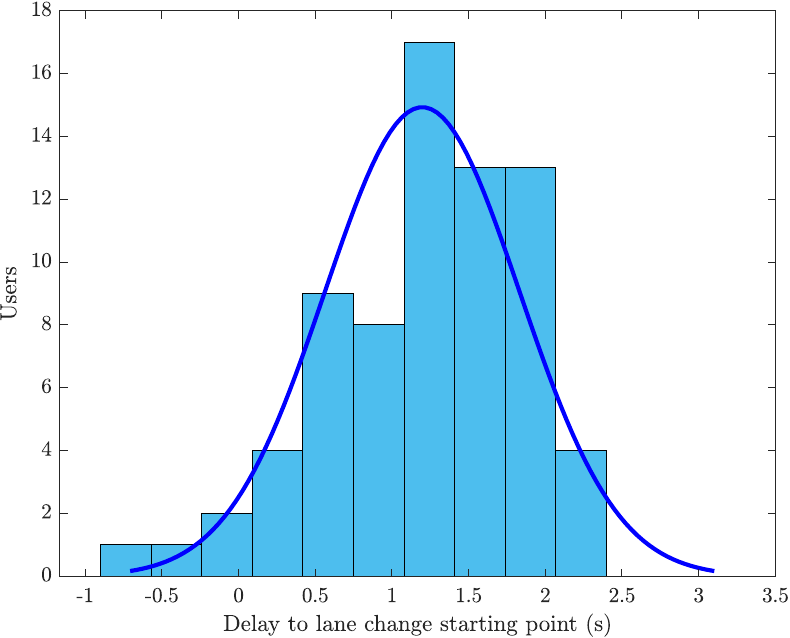}
    }
    \vspace{-18pt}
    \caption{Delay to lane-change maneuver starting point.}
    \vspace{-12pt}
    \label{fig:delayfutof0}
\end{figure}

\subsubsection{Delay to lane-change starting point}

The delay of users to lane change starting point has been defined as the difference between the frame where the user decided that the maneuver was starting and the actual frame \texttt{f0}. A histogram of users regarding the delay is observed in Figure \ref{fig:delayfutof0}, including the fit of a normal distribution ($\mu = 1.198, \sigma = 0.634$)
As regard, the mean is above 1.19 seconds; the majority of users detect lane changes after they have already started when the vehicle is clearly changing its trajectory.

\subsubsection{Anticipation to lane change middle point}

\begin{figure}[ht!]
    \centering
    {
    \includegraphics[width=\columnwidth]{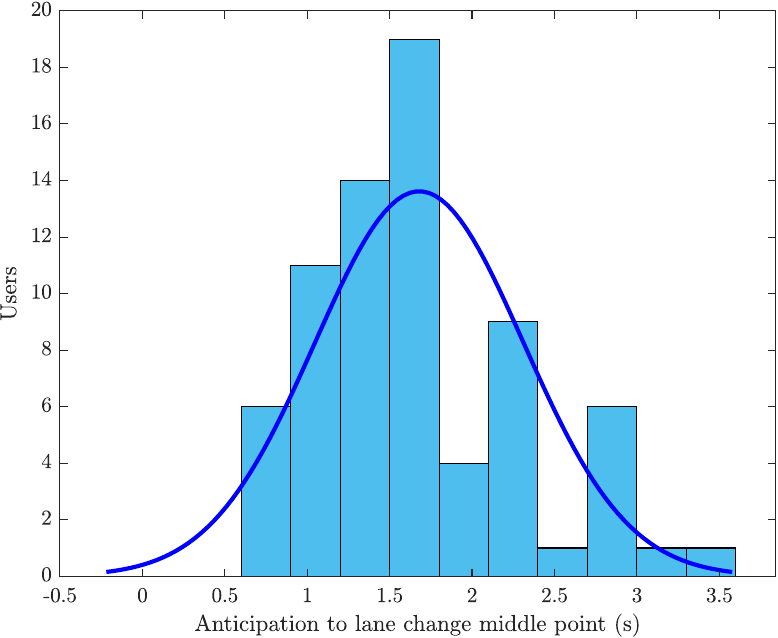}
    }
    \vspace{-16pt}
    \caption{Anticipation to lane-change maneuver middle point.}
    \vspace{-7pt}
    \label{fig:anticipationf1tofu}
\end{figure}

The anticipation of users to lane change middle point has been defined as the difference between the frame where the user decided that the maneuver was starting and the actual frame \texttt{f1}. A histogram of users regarding the anticipation is observed in Figure \ref{fig:anticipationf1tofu}, including the fit of a normal distribution ($\mu = 1.680, \sigma = 0.632$). As displayed, the users could at least detect the lane-change maneuver within a reasonable time.

\subsubsection{Delay-accuracy ratio}

\begin{figure}[t]
    \centering
    {
    \vspace{5pt}
    \includegraphics[width=\columnwidth]{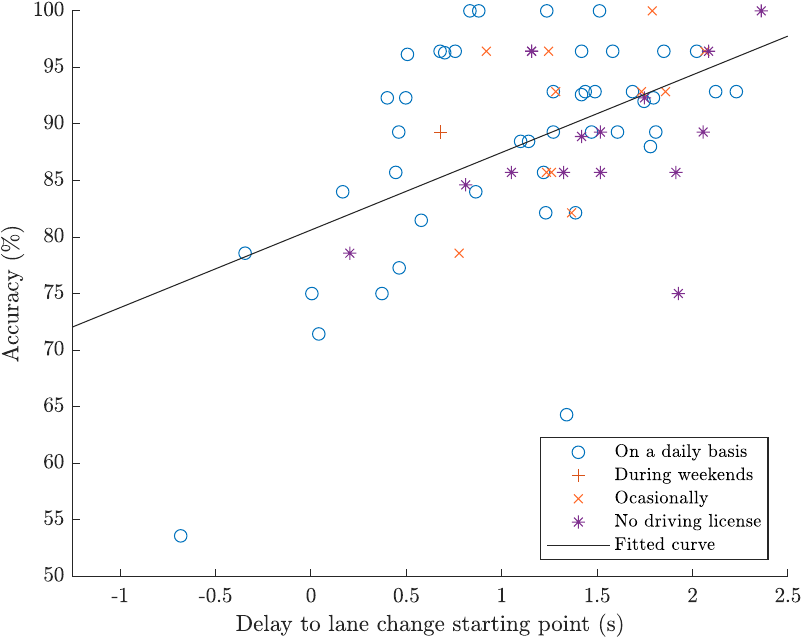}
    }
    \vspace{-15pt}
    \caption{Delay-Accuracy ratio of participants $(r = 6.86x + 80.62)$.}
    \vspace{-4pt}
    \label{fig:delay-accuracy-ratio}
\end{figure}

To extract a conclusion it is necessary to analyze the ratio between the delay to lane change starting point and accuracy. Intuitively, human beings tend to challenge themselves to improve their reaction reflex at the expense of lowering accuracy. That behavior is validated in this study, as showed in Figure \ref{fig:delay-accuracy-ratio}. Users that are more conservative obtain better accuracy, while those that score longer anticipation time tend to miss-answer in some of the sequences. 

\subsubsection{Misses}
 Errors on detection are mainly caused by thinking too far ahead in NLC sequences. When the target vehicle moves a little within the lane, the user tends to hit the button, labeling it as a LLC/RLC, leading up to an average of 1.85 errors per user (SD = 1.59).
 In LLC sequences, the error average is 0.60 (SD = 0.78), the same as on RLC sequences is 0.60 (SD = 1.02).
\subsection{Participants}
\begin{figure}[t]
    \centering
    {
    \includegraphics[width=\columnwidth]{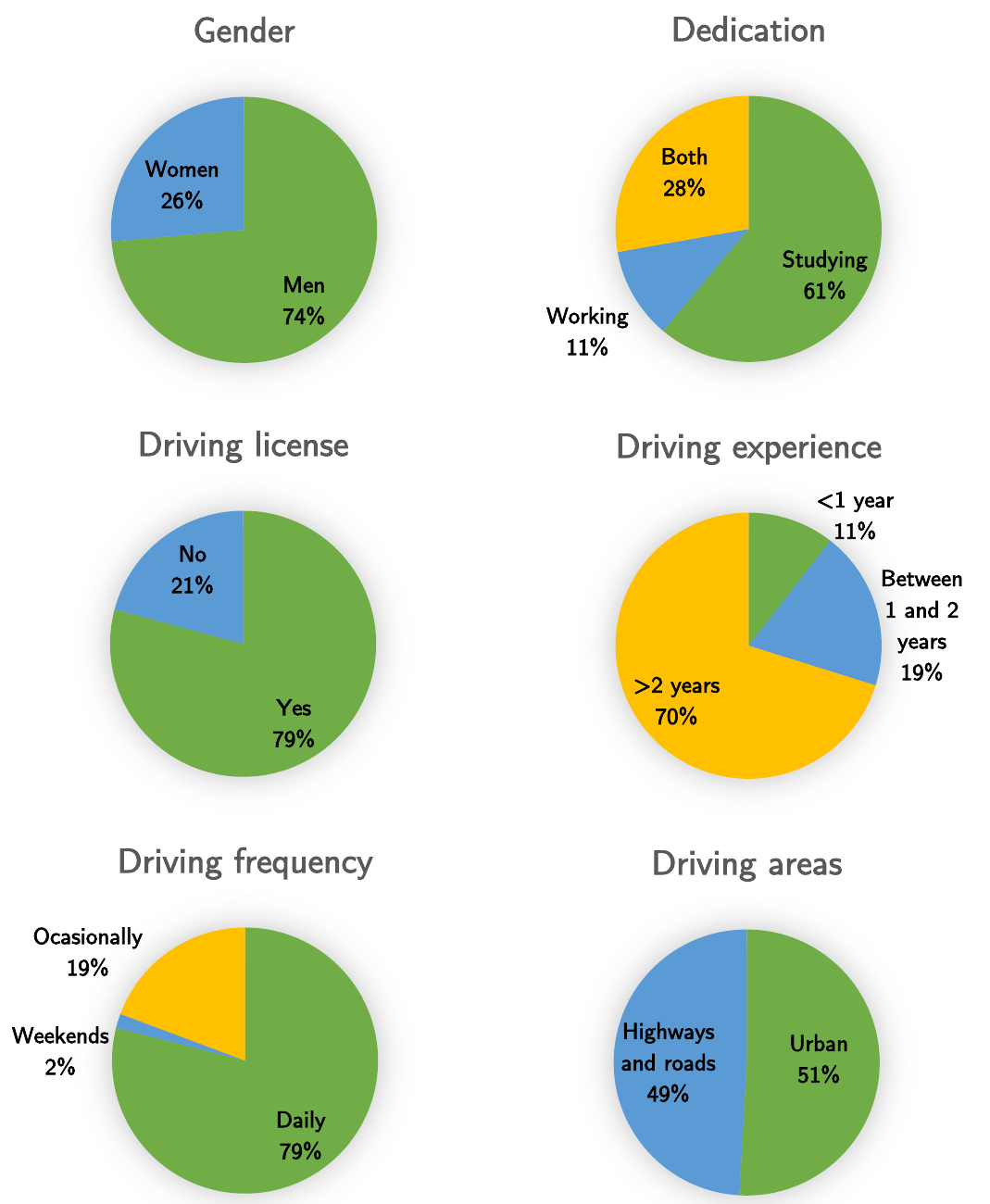}
    }
    \vspace{-19pt}
    \caption{Pie charts of users' data.}
    \vspace{-16pt}
    \label{fig:stats_graph}
\end{figure}

A total of 72 participants completed the survey, watching every selected sequence (30 out of 794). The average age of the participants was 24.22 years (SD = 10.08), and 74\% of them were male. Most of them had a driving license (79\%), and a 49\% were daily drivers in highway scenarios. In Figure \ref{fig:stats_graph} the stats about the participants are detailed.

\subsection{Statistical analysis}


Null hypothesis states that the mean of delay to lane change starting point is 0, that is, participants can detect maneuvers precisely without any lag.

Performing a Student's t-test with a significance value of 1\%, the null hypothesis is rejected, with a p-value of $p = \num{5.1615e-70}$ when testing that mean is not 0 and $p=\num{1.4184e-25}$ when testing as the alternative hypothesis that the mean is greater than 0. 
Therefore, the analysis revealed that \textbf{humans cannot anticipate a lane-change maneuver systematically}.


\section{Discussion} \label{sec:discussion}


\subsection{Trial content and design} \label{sec:trialContentDiscussion}
The survey was performed by researchers, looking for volunteers in the Polytechnic School where the INVETT Research Group is based. Moreover, professors and research colleagues were asked to take part in the Challenge, as well as students, and other assistant staff, along with individuals related to the researchers (family, friends). The design of the Challenge has been limited to the resources available and keeping in mind the nature of personal supervision needed in every single trial. 
As said before, the maneuvers are slightly imbalanced (table \ref{table:ManeuversCount}), being 47.61\% of them RLC, 29.85\% LLC, and 22.54\% NLC. Nevertheless, it was determined to introduce the data in the same proportions for each test (33.33\%) to make it impartial for the users. The dataset offers a reasonable amount of variability of all kinds of sequences, so considering the number of users and the random pick of them to create every single trial, this decision is adequate.


\subsection{User groups}
Risk-taking behavior has been analyzed within groups, being men the group that takes a more significant risk when making predictions of lane-change maneuvers. That is, they anticipate maneuvers better, but lowering the accuracy.
As shown in Figure \ref{fig:delay-accuracy-ratio}, participants that do not drive (i.e., no driving license) obtain the worst results in terms of delay and accuracy. Regarding driving experience, users whose experience is more than two years obtain the best results, especially in accuracy. Novice drivers score worse results than them. Considering driving areas, volunteers that are used to drive in highway scenarios outperformed the ones that are used to urban areas.
The conclusions obtained in this analysis are consistent with the \textit{a priori} judgment, especially in aspects such as driving experience and driving areas.




\section{Conclusions and Future Work} \label{sec:conclusions}
72 participants took part in the study, trying to predict the direction of travel of a vehicle in 30 different sequences, selected from a set of up to 800. The vehicle selected in each sequence would perform a left lane change, a right lane change, or stay in the lane. On an overall level, participants were not able to predict the direction of the vehicle, compared with GT data. Certain factors, such as front wheels angling, merging maneuvers, or the use of turn signal increased participants' reliability, making them predict better the direction in those situations, as been qualitative observed. 

As future work, the main task would be to improve the Challenge, adding cue labeling and extending the number of participants, even using an external survey participation service, such as Amazon Mechanical Turk. Adding more experienced drivers to the study would improve prediction correctness and certainty, which could lead to compare it to advanced prediction algorithms, such as \cite{RubenITSC19}, in order to develop better architectures and understand what is the best approach to predict lane changes successfully.
Another exciting research work would be to monitor participants' eyes and faces with a frontal camera while completing the survey and analyze their reactions and cue extracting abilities.
Finally, a future approach could be developed by using a simulator (e.g., CARLA \cite{CARLA}) along with an immersing cockpit that could assess users' capacity to avoid hazardous situations that they can spot while driving, being previously programmed in several trials regarding tailgating maneuvers, avoidance of rear-end collisions and evasive maneuvers when driving into an obstacle on the road.

\section*{ACKNOWLEDGMENT}
This work was funded by Research Grants S2018/EMT-4362  (Community Reg. Madrid), DPI2017-90035-R  (Spanish Min. of Science and Innovation), BRAVE Project, H2020, Contract \#723021 and PRE2018-084256 (Spanish Min. of Education) via a predoctoral grant to the first author. It has also received funding from the Electronic Component Systems for European Leadership Joint Undertaking under grant agreement No 737469 (AutoDrive Project). This Joint Undertaking receives support from the European Unions Horizon 2020 research and innovation programme and Germany, Austria, Spain, Italy, Latvia, Belgium, Netherlands, Sweden, Finland, Lithuania, Czech Republic, Romania, Norway.



\bibliographystyle{bibliography/IEEEtran}
\bibliography{bibliography/IEEEabrv,bibliography/references}

\end{document}